\documentclass{article}

\usepackage{arxiv}

\usepackage[utf8]{inputenc} 
\usepackage[T1]{fontenc}    
\usepackage{hyperref}       
\usepackage{url}            
\usepackage{booktabs}       
\usepackage{amsfonts}       
\usepackage{nicefrac}       
\usepackage{microtype}      
\usepackage{lipsum}		
\usepackage{graphicx}
\usepackage{natbib}
\usepackage{doi}
\usepackage{subfig}  
\usepackage{multirow}
\usepackage{caption}
\usepackage[blocks]{authblk}

\usepackage[final]{changes}

\title{\textit{TASTEset} - Recipe Dataset and Food Entities Recognition Benchmark}

\date{} 					

\author[1,$\dag$]{\href{https://orcid.org/0000-0002-3407-7570}{Anna Wróblewska}}

\author[2,$\dag$]{\href{https://orcid.org/0000-0003-2856-8901}{Agnieszka Kaliska}}

\author[3,$\dag$]{\href{https://orcid.org/0000-0002-3096-9918}{Maciej Pawłowski}}

\author[4]{\href{https://orcid.org/0000-0003-1194-7921}{Dawid Wiśniewski}}

\author[1]{\href{https://orcid.org/0000-0002-2241-9588}{Witold Sosnowski}}

\author[4,5,$\dag$]{\href{https://orcid.org/0000-0002-2442-345X}{Agnieszka Ławrynowicz}}

\affil[$\dag$]{EQUAL CONTRIBUTION}

\affil[1]{Faculty of Mathematics and Information Science, Warsaw University of Technology, Warsaw, Poland\\ 
\texttt{\{anna.wroblewska1,witold.sosnowski.dokt\}@pw.edu.pl}}

\affil[2]{Faculty of Modern Languages, Adam Mickiewicz University, Poznan, Poland \\
\texttt{akaliska@amu.edu.pl}}

\affil[3]{\texttt{mac.pawlowski19@gmail.com}}
\affil[4]{Faculty of Computing and Telecommunications 
	\\ Poznan University of Technology, Poznan, Poland\\
	\texttt{\{dawid.wisniewski,alawrynowicz\}@put.poznan.pl}}

\affil[5]{Center for Artificial Intelligence and Machine Learning (CAMIL)\\ Poznan University of Technology, Poznan, Poland}

\renewcommand{\shorttitle}{\textit{TASTEset} - Recipe Dataset and Food Entities Recognition Benchmark}

\hypersetup{
pdftitle={TASTEset - Recipe Dataset and Food Entities Recognition Benchmark},
pdfsubject={cs.AI,cs.CL,cs.IR,cs.LG},
pdfauthor={Ania Wróblewska, Agnieszka Kaliska, Maciej Pawłowski, Dawid Wiśniewski, Witold Sosnowski, Agnieszka Ławrynowicz},
pdfkeywords={BioNLP Application, Food Computing, Named Entity Recognition},
}

\begin{document}
\maketitle

\begin{abstract}

Food Computing is currently a fast-growing field of research. Natural language processing (NLP) is also increasingly essential in this field, especially for recognising food entities.  

However, there are still only a few well-defined tasks that serve as benchmarks for solutions in this area. We introduce a new dataset -- called \textit{TASTEset} -- to bridge this gap. In this dataset, Named Entity Recognition (NER) models are expected to find or infer various types of entities helpful in processing recipes, e.g.~food products, quantities and their units, names of cooking processes, physical quality of ingredients, their purpose, taste. 

The dataset consists of 700 recipes with more than 13,000 entities to extract. We provide a few state-of-the-art baselines of named entity recognition models, which show that our dataset poses a solid challenge to existing models. The best model achieved, on average, 0.95 $F_1$ score, depending on the entity type -- from 0.781 to 0.982. We share the dataset and the task to encourage progress on more in-depth and complex information extraction from recipes. 
\end{abstract}


\section{Introduction}

Food computing is a multidisciplinary field as it integrates food, nutritional, and computer sciences, as well as gastronomy~\cite{food-computing-survey}. Preparing a health-aware food management system requires a vast amount of domain knowledge and a deep understanding of recipes, especially extracting crucial information. This research is a step toward deeper analysing and extracting vital information from recipes.

The main contributions of this study are:
\begin{enumerate}
\item \textit{TASTEset} -- our novel dataset of 700 text recipes with 3,788 ingredients and above 13,000 entities of different types (helpful in food computing);
\item Our method of collecting and annotating the dataset (see Section~\ref{sec:dataset}); 
\item Evaluation of several state-of-the-art Named Entity Recognition (NER) baselines (see Sections~\ref{sec:experiments} and~\ref{sec:results});
\end{enumerate}

In the following, Section~\ref{sec:related-work} describes our motivation and related work. Then, we describe our approach to preparing the dataset and perform exploratory dataset analysis in Section~\ref{sec:dataset}. We also present experiments with the state-of-the-art baselines (Section~\ref{sec:experiments}) and their results (Section~\ref{sec:results}). Finally, we conclude our paper in Section~\ref{sec:conclusions}.

\section{Related Work}\label{sec:related-work}

Our main reason for preparing a new dataset was to develop a strategy to deal with challenges faced by food computing, especially concerning a deeper understanding of information about ingredients in recipes, e.g.~their quantities, processing, and various aspects regarding different dietary needs. 
This comprehensive information then can advise proper nutrition, e.g. it can be utilized to build health-aware recommendation systems.

According to the survey in~\cite{food-computing-survey}, existing datasets and methods are not well prepared for health-aware and lifestyle preferences or constraints. In order to design health-aware systems for recipe recognition and recommendations, we need to understand recipes deeply, e.g.~recognising specific ingredients, their groups, understanding their relations and functions, quantities, or other dietary issues. 
For example, different degrees of food or ingredient processing show whether something is healthy or not~\cite{Menichetti2021.05.22.21257615}. In order to prepare more advanced analysis or recommendations considering all food processing aspects, ingredient quantities, etc., we need fine-grained as well as very precise entity extraction and analysis.

However, these knowledge and notation is scarce in currently available datasets to train machine learning models. 
Available recipe datasets are mainly focused on information retrieval (mainly visual or cross-modal), image-text recipe generation, or extracting general information about cooking and ingredients. \cite{food-computing-survey} survey existing benchmark food datasets and challenges. 
The most extensive recipe datasets are Recipe1M+ ~\cite{recipe1m} and RecipeNLG~\cite{DBLP:conf/inlg/BienGMTWL20}. However, they lack specific fine-grained tags related to ingredients. 

Though there is a substantial number of available annotated datasets with entities from the biomedical domain, such resources are still insufficient in the food domain. 
The available corpora include r-FG (recipe flow graph)~\cite{mori2014flow}, CURD (Carnegie Mellon University Recipe Database)~\cite{de2008guide}, and FoodBase~\cite{popovski_foodbase_2019}.  
The r-FG dataset is composed of 266 Japanese recipes, annotated with eight tags, associated with food, tool, duration, quantity, chef's action, action by foods, food state, and tool state.
CURD consists of 300 annotated and 350 unannotated recipes, for which the Minimal Instruction Language for the Kitchen Language (MILK) is used.
FoodBase~\cite{popovski_foodbase_2019} comprises of 12,844 food entity annotations describing 2,105 food entities, which are later linked to the Hansard corpus~(\url{https://www.hansard-corpus.org}). 
All the above-mentioned corpora are limited. The r-FG dataset consists of only Japanese food recipes. 
All the state-of-art corpora use annotations schemes that have limitations, in particular they are \emph{too coarse-grained}, mostly providing only general food entities.

The goal of making entities recognition automatic can be achieved by applying natural language processing methods (i.e., Named Entity Recognition (NER)~\cite{DBLP:journals/csur/NasarJM21}). 
Classical approaches for Named Entity Recognition frequently involved Conditional Random Fields (CRFs) proposed by Lafferty et al.~\cite{DBLP:conf/icml/LaffertyMP01}. Currently, most of the methods solving NER problems use deep neural networks to capture context information about a given token. Huang et al.~\cite{DBLP:journals/corr/HuangXY15} proposed a bidirectional recurrent neural network using long short-term memory (LSTM) cells~\cite{hochreiter1997long} enriched with a CRF layer to tag sequences with predefined labels. In recent years, the focus shifted from recurrent neural networks to attention-based models that allow large-scale parallel training and overcome problems with long dependencies observed among recurrent architectures. Devlin et al.~\cite{devlin-etal-2019-bert} proposed BERT, a bidirectional encoder representation based on the Transformer architecture~\cite{DBLP:conf/nips/VaswaniSPUJGKP17}. BERT was shown as an efficient model that may be used to recognize named entities. Yamada et al. proposed LUKE~\cite{yamada-etal-2020-luke}, a method based on Transformer architecture that is trained using a new pretraining task based on BERT. Akbik et al.~\cite{DBLP:conf/coling/AkbikBV18} proposed contextual string embeddings based on character-level language models that provide information to bidirectional LSTM with CRF layers.

In~\cite{make4010012}, the authors indicate biomedical named entity recognition (bioNER) as a sub-field of NER approaches. The bioNER field copes with biological entities, such as genes, diseases, or names of biological substances, e.g. food names. This survey divides bioNER techniques into rule-based models, dictionary-based models, machine learning-based models, and hybrid models. The authors identified food dedicated solutions: BuTTER method based on a bidirectional long-short term memory (LSTM)~\cite{cenikj_butter_2020}, FoodIE rule-based detector~\cite{popovski_foodie:_2019}, and ontology-based NCBO annotator. This NCBO annotator can be used with any ontology, particularly also with ontology containing concepts associated with food and dietary entities~\cite{jonquet2009ncbo}. It works as a dictionary-based technique limited to only simple data preprocessing, so it can only recognize entities its dictionaries/ontologies contain~\cite{popovski_survey_2020}. The authors of~\cite{make4010012} also provided extensive tests and compared their hybrid dictionary and CRF-based model, FoodCoNER, that outperforms the other bioNER models and four deep language models: BERT, BioBERT, RoBERTa, and ELECTRA. The results, yet promising, were made on the FoodBase dataset containing only food product entities.









\section{Dataset} \label{sec:dataset}
A total of 700 sets of ingredients from recipes were manually annotated using BRAT annotation tool~\cite{stenetorp-etal-2012-brat}.
The recipes were scraped from the websites \url{https://www.allrecipes.com/}, \url{http://food.com/}, \url{https://tasty.co/},  and \url{https://www.yummly.com/}, using the tool \url{https://github.com/hhursev/recipe-scrapers}. 
In this paper, we discuss 9 entity types. 

\subsection{Tagset Description}
\begin{figure}[!htbp]
    \centering
    \subfloat[PROCESS entity]{\includegraphics[width=0.7\textwidth]{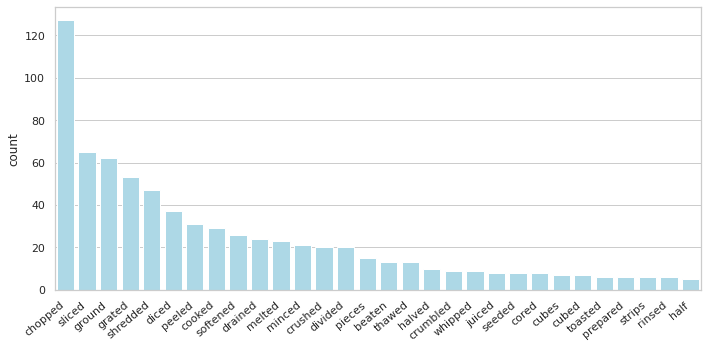}
    }
    \hfill
    \subfloat[PHYSICAL QUALITY entity]{\includegraphics[width=0.7\textwidth]{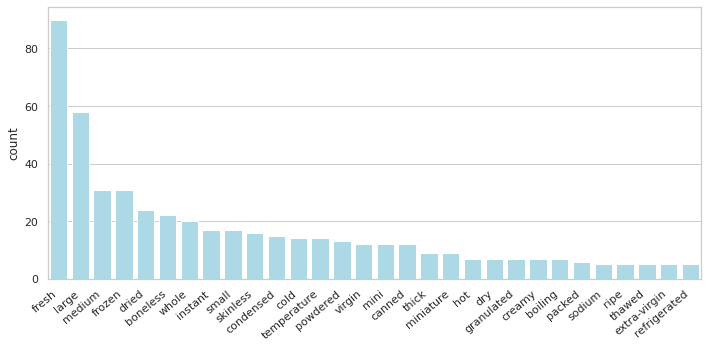}}
    \caption{The most frequent values of the PROCESS and PHYSICAL QUALITY entity types}
    \label{fig:process_and_physical_quality_pie_charts}
\end{figure}

The manual annotation covered 3,788 ingredients of varying complexity. These ranged from relatively simple ingredients (e.g.~\textit{2 eggs}) to much more complex ones (e.g.~\textit{2 lbs sweet potatoes, peeled and cut into 1/8 inch thick slices}), which revealed that the name of an ingredient is often accompanied by a variety of its attributes. On this basis, we identified entities such as:
\begin{itemize}
    \item FOOD as the name of an ingredient (e.g.~\textit{bread}, \textit{mayonnaise}, \textit{salt}, \textit{tomato}),
    \item QUANTITY as a quantity (usually expressed by digits or a float),
    \item UNIT as a unit of measurement (e.g.~\textit{bunch}, \textit{cup}, \textit{grams}, \textit{jar}, \textit{millimeters}, \textit{slices}, \textit{stalks}, \textit{tablespoon}, \textit{teaspoon}, \textit{ounces}, \textit{package}, \textit{pinch})
    \item PROCESS as the attribute of the ingredient, usually referring to an action to be taken to prepare the ingredient (e.g.~\textit{chopped} for parsley, \textit{crushed} for garlic, \textit{grated} for ginger, \textit{ground} for black pepper, \textit{minced} for garlic, \textit{quartered} for onion, \textit{softened} for butter, \textit{shredded} for cheese),
    \item PHYSICAL QUALITY as the characteristic of the ingredient (e.g.~\textit{boneless} for chicken breast, \textit{frozen} for spinach, \textit{fresh} or \textit{dried} for basil, \textit{powdered} for sugar),
    \item COLOR as the color of the ingredient (e.g.~\textit{brown} for rice, \textit{green} for apples, \textit{white} for bread),
    \item TASTE as the flavour (e.g. \textit{bittersweet}, \textit{butter-flavoured}, \textit{sweet}, \textit{semi-sweet}),
    \item PURPOSE as the purpose of using the ingredient in the recipe (e.g.~\textit{for dusting} about flour, \textit{for garnish} about sunflower seeds, \textit{for frying} about canola oil, and \textit{as topping} about sour cream),
    \item PART as a part of the ingredient required by the recipe (e.g.~\textit{breast} and \textit{wing} as parts of the chicken, \textit{heart} as part of the artichoke, \textit{yolks} and \textit{whites} as parts of the eggs).
\end{itemize}
As one can easily guess, some entities appear in the corpus more frequently (e.g. UNIT and QUANTITY), while others are less frequent (e.g. COLOR and TASTE). And within each category, there will be both very frequent and very rare entities - see Figure~\ref{fig:process_and_physical_quality_pie_charts} for the percentage ratios of the ten most frequent attributes of the PROCESS and PHYSICAL QUALITY entity types as an example.

\subsection{Tags and Values: Common and Less Common Entity Examples}
While it was possible to assume, for example, that entities of the PROCESS type would be typical past participles ending in \textit{-ed}, it soon turned out that either not every participle should be treated as a process name (e.g.~\textit{salted} is a TASTE in the case of butter and \textit{granulated} is a PHYSICAL QUALITY in the case of sugar), or the PROCESS tag took on more complex values -- cf.~e.g.~\textit{thinly sliced} in the example \textit{3 cups onion (thinly sliced)}, \textit{cooked al dente} in \textit{2 cups pasta cooked al dente}, \textit{cut in small wedges} in \textit{1 head cabbage cut in small wedges} and \textit{cut lengthwise in half} (about peppers).

\begin{figure}[!htbp]
    \centering
    \includegraphics[width=0.45\textwidth]{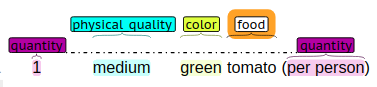}
    \caption{Quantity as a discontinuous value (corpus example)
    }\label{fig:taisti1-food}
\end{figure}

Almost every entity listed above has taken on quite unusual, context-dependent values in addition to the expected ones, e.g.~\textit{regular} is a COLOR in the example \textit{15 ounces oreos (white or regular)}, \textit{bone-in} and \textit{with skin} are PHYSICAL QUALITIES of the chicken, and \textit{1 per person} indicates the QUANTITY of tomatoes - the entity is discontinuous - see Figure~\ref{fig:taisti1-food}.

\subsection{Proper Names}\label{section:proper_names}
The names of ingredients are mostly simple \textit{apellativa}. However, in the corpus we found quite a few examples of proper names written with a capital letter, naming products of specific brands or producers (e.g.~\textit{Jello} for a gelatine producer, \textit{Splenda} for a sweetener producer, as well as more complex product names, e.g.~\textit{NESTLÉ TOLL HOUSE Refrigerated Sugar Cookie Bar Dough}), and besides them quite a few examples of proper names written with a lower case letter (e.g.~\textit{breyers} for a producer of chocolate ice-cream, \textit{truvia} for a producer of sweeteners). There are also quite a few examples of names written with a capital letter, which in a way are proper names, but they do not really refer to brand names, but to various kinds of products produced today in different parts of the world (e.g. \textit{Dijon} for mustard, \textit{Gouda}, \textit{Gruyère}, \textit{Monterey Jack} and \textit{Parmesan} for cheese types, \textit{Tabasco} and \textit{Worcestershire} for sauces, \textit{Bourbon}, \textit{Cointreau}, \textit{Kahlua} and \textit{Triple Sec} for alcohols and liqueurs).

\subsection{Distribution of Ingredients}
\begin{figure}[!htbp]
    \centering
    \includegraphics[width=0.18\textwidth]{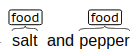}
    \caption{Two ingredients on one line without indication of measures and quantities (simple case).
    }\label{fig:taisti2-food}
\end{figure}

\begin{figure}[!htbp]
    \centering
    \includegraphics[width=0.3\textwidth]{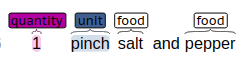}
    \caption{Two ingredients on one line with common measure and quantity (complex case).
    }\label{fig:taisti3-food}
\end{figure}

\begin{figure}[!htbp]
    \centering
    \includegraphics[width=0.42\textwidth]{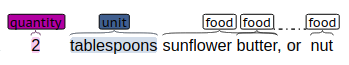}
    \caption{Two ingredients on one line with common measure and quantity (alternative ingredients).
    }\label{fig:taisti4-food}
\end{figure}

While generally one ingredient is indicated on a single line in the ingredient set, it is quite common for another ingredient, additional or alternative, to be indicated explicitly or implicitly in the same line, next to the \textit{main} ingredient (e.g.~\textit{salt and pepper} vs. \textit{2 tablespoons sunflower butter, or nut} vs. \textit{1 1⁄2 cups heavy cream, plus additional, lightly whipped, for serving}). Of these three examples, \textit{nut butter} and \textit{(additional) heavy cream} are discontinuous. Manual annotation was done in BRAT, which allows for the annotation of such cases.\footnote{https://brat.nlplab.org} The colors below refer to different entities. As can be seen in the Figure~\ref{fig:taisti2-food}, Figure~\ref{fig:taisti3-food} and Figure~\ref{fig:taisti4-food}, where two ingredients were indicated on one line, usually the quantity and measure were indicated only once, referring to both the first indicated ingredient and the second indicated ingredient, which was then some distance away from the quantity and/or measure. 

It is worth pointing out, in addition to the above-mentioned characteristics, that the corpus we manually annotated was rather noisy, containing repetitions and spelling errors and, besides the names of ingredients, their quantities and measures, also additional comments (e.g. on the availability of a given product in such and such a shop for such and such a price). These features occurred irregularly in the corpus, with varying degrees of intensity, making it at the same time very challenging.

\subsection{Exploratory Analysis}\label{sec:eda1}

The \textit{TASTEset} comprises 700 recipes with an average number of ingredients of 5.41 (from 1 to 7 ingredients). Table~\ref{tab:entity-statistics} lists the detailed statistics of the entities in the dataset.

\begin{table}[!htbp]
    \footnotesize
	\centering
	\begin{tabular}{lllllll}
		\toprule
		Entity & Total  & Unique  & Avg. entity &  \% & Example & Avg. token  \\
		 &  count &  values & count/recipe &  in the set &  gold value &  count/entity \\
		\midrule
		FOOD &  4,020 & 1,123 & 5.74 & 30.08  & salt & 1.52\\
		QUANTITY &  3,780 & 206 & 5.4 & 28.29 & 1/2 & 1.10\\ 
		UNIT &  3,172 & 127 & 4.53 & 23.74 & teaspoon & 1.01\\ 
		PROCESS & 1,091 & 254 & 1.56 & 8.16 & chopped & 1.44\\
		PHYSICAL QUALITY & 793 & 202 & 1.13 & 5.93 & fresh & 1.24\\
		COLOR & 231 & 24 & 0.33 & 1.73 & white & 1.02\\
		TASTE &  126 & 31 & 0.18 & 0.94 & sweet & 1.06\\
		PURPOSE & 94 & 33 & 0.13 & 0.70 & garnish & 2.23\\
		PART &  55 & 18 & 0.08 & 0.41 & whites & 1.16\\
		\midrule
		all &  13,362 & 1,969 & 19.09 & 100 & -- & 1.25\\
		\bottomrule
	\end{tabular}
	\caption{Summary of the entities in the \textit{TASTEset} dataset\label{tab:entity-statistics}}
\end{table}

\begin{figure*}[!htbp]
\centering
\includegraphics[width=.7\textwidth]{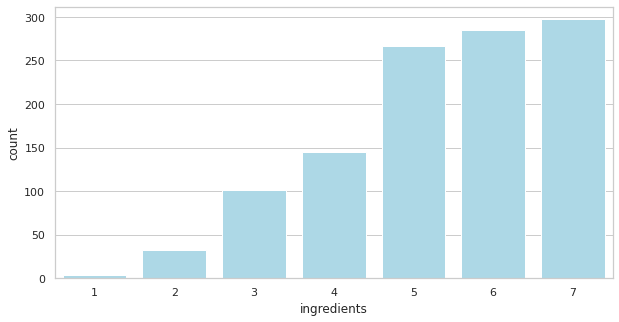}\hfill
\caption{The distribution of the entities in the \textit{TASTEset} dataset}
\label{fig:ent_dist}
\end{figure*}

The majority of the entities in the dataset belong to FOOD, QUANTITY and UNIT while the least represented entities are TASTE, PURPOSE and PART. Although such an imbalance of entities in a dataset seems natural in the field of recipes, it must be taken into account when building machine learning models. Figure~\ref{fig:ent_dist} shows the entity distribution.

\section{Experiments} \label{sec:experiments}

The \textit{TASTEset} dataset is also challenging because of the fine-grained structure of entities and, from time to time, different semantic meanings of the same tokens, e.g. \textit{garlic} can be a taste or a food entity, \textit{hot} can be a taste (e.g. \textit{hot sauce}), a physical quality (e.g. \textit{hot water}) or a part of a food name (e.g. \textit{hot dog}). The annotation of some elements describing ingredients is problematic. It may be a matter of interpretation, e.g. \textit{egg noodles} can be interpreted as a part of the food entity in the list of ingredients or as a physical quality of it: \textit{8 ounces dried pasta (egg noodles or fettuccine)}. Although in most cases, the interpretation seems obvious as in example \textit{8 ounces fine egg noodles} where \textit{egg noodles} is a food name.

\begin{figure}[!htbp]
  \centering
  \begin{tabular}[c]{l}
      List of ingredients \\\hline
      - 1 box powdered sugar \\
      - 8 oz. soft butter \\   
      - 1 (8 oz.) peanut butter \\   
      - paraffin \\   
      - 12 oz. chocolate chips \\ 
  \end{tabular} \qquad
  \begin{tabular}[c]{c}
    \includegraphics[width=.47\linewidth]{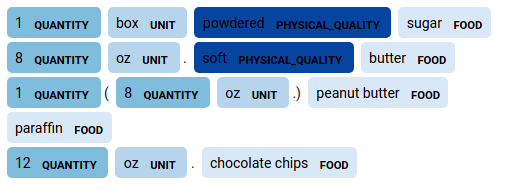} \\
  \end{tabular}
   \caption{Example of a model predictions -- recognition of entities in one recipe from \textit{TASTEset} dataset.
   \label{fig:model-prediction}}
\end{figure}

The goal of the experiments was to establish baselines for properly indicating entities in a recipe text (i.e. in its list of ingredients) and recognizing its semantic meaning as one from the tag list: FOOD, UNIT, QUANTITY, PHYSICAL QUALITY, PROCESS, COLOR, TASTE, PURPOSE, PART. 
Figure~\ref{fig:model-prediction} presents an example of predictions achieved by one of our baselines.

We trained a set of NER models using one of the state-of-the-art architectures working on plain texts. We focused on BERT~\cite{devlin-etal-2019-bert} and LUKE~\cite{yamada-etal-2020-luke}.

As for the BERT, we tested three models: BERT\texorpdfstring{\textsubscript{base-cased}}{}, BERT\texorpdfstring{\textsubscript{large-cased}}{}, FoodNER. The first two were released by BERT's authors~\cite{devlin_bert:_2019}. FoodNER~\cite{stojanov_fine-tuned_2021} is a BERT with token classification layer, and is already fine-tuned on the food entity recognition problem, on the FoodBase dataset~\cite{popovski_foodbase_2019}, and might contain domain-specific knowledge. However, \textit{TASTEset} has more entities, therefore only the BERT part from FoodNER could be used. The classification layer was trained once again from the beginning.

We tested two types of architectures that utilized BERT. Namely, BERT with a dropout and a classification layer~\footnote{\url{https://huggingface.co/docs/transformers/main/en/model\_doc/bert\#transformers.BertForTokenClassification}}, and the second which additionally had a CRF layer on top of it. The CRF layer exploits the dependencies between consecutive entities while inference. In \textit{TASTEset}, the entities are naturally highly codependent, for example the QUANTITY entity is usually followed by UNIT (e.g. \textit{1 teaspoon}), or COLOR and PHYSICAL QUALITY usually directly precede or follow the FOOD entity (e.g. \textit{brown sugar}).  For the BERT with CRF model implementation we closely followed~\cite{souza2020bertimbau}.~\footnote{\url{https://github.com/neuralmind-ai/portuguese-bert/tree/master/ner\_evaluation}}

For all BERT-related models, the single input consisted of at most 128 tokens. We used the AdamW optimizer with linear scheduler, and weight decay equal to $0.01$. The mini batch size was set to 16. We used 0.2 as the dropout rate in the classification layer. The following parameters were optimized with the grid search: number of training epochs -- \{10, 15, 20, 25, 30\}, the learning rate -- \{$2\cdot10^{-5}$, $5\cdot10^{-6}$\}. Table~\ref{tab:baseline-hyperparameters}  presents the hyperparameters that yielded the best results during the cross-validation. For the BERT + CRF results, we depicted only the ones associated with BERT-large-cased, as this model was the best in the basic scenario.

\begin{figure*}[h!]
\centering
\includegraphics[width=.6\textwidth]{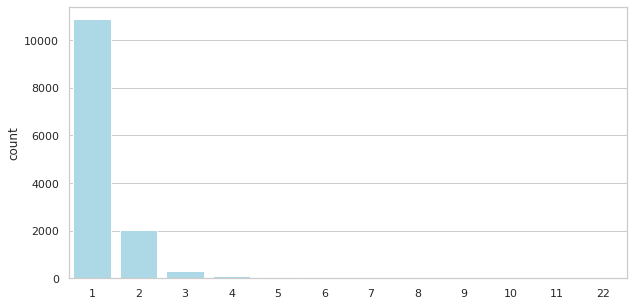}\hfill
\caption{Entity length distribution (number of tokens in entities).}
\label{fig:entity_length_distribution}
\end{figure*}

We tested LUKE model from huggingface library\footnote{\url{https://huggingface.co/docs/transformers/model_doc/luke}} fine-tuned according to the official huggingface script \footnote{\url{https://github.com/huggingface/transformers/tree/main/examples/research_projects/luke}} based on the pre-trained model "studio-ousia/luke-base"~\cite{yamada-etal-2020-luke}.\footnote{model "studio-ousia/luke-base" is available at \url{https://huggingface.co/studio-ousia/luke-base}} Luke-base is pretrained on RoBERTa-base which is case-sensitive.
In our experiment, the LUKE head used for the testing was \textit{LukeForEntitySpanClassification}.  We used the AdamW optimizer with linear scheduler, and weight decay equal to $0.0$.
The best hyperparameters set was obtained during 5-fold cross-validation. The following parameters were optimized: number of training epochs -- \{10, 15, 20, 25, 30\}, $\textrm{mini\_batch\_size}$ -- \{8, 16\}, and LUKE-specific parameters: $\textrm{max\_mention\_length}$ (the maximum number of tokens inside an entity span) -- \{3, 5, 30\} (see Figure~\ref{fig:entity_length_distribution}), $\textrm{max\_entity\_length}$ (the maximum total input entity length after tokenization) -- \{32, 128, 512\}.

At the evaluation stage, we used the NER model to detect all entity occurrences in the text. We utilized $F_1$-score, with exact-boundary and type
matching~\cite{segura-bedmar-etal-2013-semeval}.~\footnote{We used  \textit{nervaluate} module for metrics calculation \url{https://pypi.org/project/nervaluate/}}

In our experiments, we used 5-fold cross-validation. All the results are reproducible with the seed $42$. We have not applied any additional preprocessing to the inputs, especially, we preserved the newline character, as it separates different  ingredients and their attributes from one another (see Figure~\ref{fig:model-prediction}). 

\section{Results}\label{sec:results}

Table~\ref{tab:cross-val-all-models} shows the results of baseline models for our \textit{TASTEset} datasets.\footnote{The source code and models' details are available at \url{https://github.com/taisti/TASTEset}.}
The baselines we utilized were models being one of the best, or even state-of-the-art in standard NER tasks. For the BERT with CRF model, we presented only the best results, which were achieved with the BERT\texorpdfstring{\textsubscript{base-cased}}{} model. The results are obtained through 5-folds cross-validation, we provided the mean score along with standard deviations of all runs. 

\begin{table}[!htbp]
    \footnotesize
	\centering
	\begin{tabular}{cccccc}
		\toprule
		& \multicolumn{5}{c}{Model} \\
		\cmidrule(r){2-6}
		entity & BERT\texorpdfstring{\textsubscript{base-cased}}{} & BERT\texorpdfstring{\textsubscript{large-cased}}{} & FoodNER & BERT with CRF & LUKE\\
		\midrule
		FOOD & $0.889 \pm 0.013$ & $0.898 \pm 0.012$ & $0.888 \pm 0.010$ & $0.895 \pm 0.013$ & $\mathbf{0.915 \pm 0.005}$\\
		QUANTITY & $0.983 \pm 0.004$ & $0.982 \pm 0.007$ & $\mathbf{0.985 \pm 0.004}$ & $0.980 \pm 0.006$ & $0.975 \pm 0.010$\\ 
		UNIT & $\mathbf{0.979 \pm 0.005}$ & $0.976 \pm 0.009$ & $\mathbf{0.979 \pm 0.005}$ & $0.975 \pm 0.006$ & $0.976 \pm 0.006$ \\ 
		PROCESS & $\mathbf{0.930 \pm 0.006}$ & $0.927 \pm 0.015$ & $0.921 \pm 0.007$ & $0.913 \pm 0.019$ & $0.916 \pm 0.021$\\
		PHYSICAL QUALITY & $0.797 \pm 0.011$ & $0.795 \pm 0.025$ & $\mathbf{0.804 \pm 0.015}$ & $0.772 \pm 0.003$ & $0.767 \pm 0.035$ \\
		COLOR & $0.873 \pm 0.039$ & $0.910 \pm 0.039$ & $\mathbf{0.913 \pm 0.037}$ & $0.894 \pm 0.046$ & $0.872 \pm 0.051$ \\
		TASTE & $0.772 \pm 0.115$ & $0.781 \pm 0.089$ & $0.769 \pm 0.116$ & $\mathbf{0.789 \pm 0.093}$ & $0.655 \pm 0.120$\\
		PURPOSE & $0.789 \pm 0.145$ & $\mathbf{0.853 \pm 0.141}$ & $0.801 \pm 0.140$ & $0.821 \pm 0.127$ & $0.684 \pm 0.176$\\
		PART & $0.705 \pm 0.115$ & $\mathbf{0.801 \pm 0.034}$ & $0.744 \pm 0.127$ & $0.730 \pm 0.091$ & $0.786 \pm 0.10$\\
		all & $0.932 \pm 0.008 $ & $\mathbf{0.935 \pm 0.011}$ & $0.932 \pm 0.006$ & $0.929 \pm 0.008$ & $0.927 \pm 0.008$\\
		\bottomrule
	\end{tabular}
	\caption{The detailed results (average $F_1$-scores over 5 runs in 5-fold cross-validation) of our baselines for \textit{TASTEset} dataset.\label{tab:cross-val-all-models}}
\end{table}

Based on the achieved results, we see that BERT-based models are better for the food entity recognition task on the \textit{TASTEset}. LUKE achieved higher scores only for the FOOD entity, with average $F_1$-score over 0.9. For QUANTITY, UNIT, PROCESS and PHYSICAL QUALITY, the difference between BERT models and LUKE is negligible. However, for the less frequent entities, LUKE performs significantly worse. 

All BERT models obtained similar performance on most of the entities. However, the BERT\texorpdfstring{\textsubscript{large-cased}}{} handles the rarest entities: PURPOSE and PART much better than all other models. For that particular reason, we believe that this is the best model out of all that we have tested in this work. We did not find adding CRF layer to the BERT\texorpdfstring{\textsubscript{large-cased}}{} to improve the model's performance. 

Table~\ref{tab:baseline-hyperparameters} depicts our baselines' customized hyperparameters to reproduce the results. All the parameters that are not mentioned were left as default, and can be derived from the models' default library settings. 

\begin{table}[!htbp]
\footnotesize
 	\centering
 	\begin{tabular}{rp{10cm}}
		\toprule
 		Training technique & Evaluated on 5-fold cross-validation using seed 42 (models' weights also initialized with seed 42). \\
 		Training data & \textit{TASTEset} \\
 		Tag dictionary & {FOOD, UNIT, QUANTITY, PHYSICAL QUALITY, PROCESS, COLOR, TASTE, PURPOSE, PART, O (not an entity)}, BERT-based models trained in BIO format, LUKE according to its own procedure: text + potential entity spans to classify\\
 		\bottomrule
 	\end{tabular}
 	\begin{tabular}{rlllll}
 		Model & BERT\texorpdfstring{\textsubscript{base-cased}}{} & BERT\texorpdfstring{\textsubscript{large-cased}}{} & FoodNer & BERT with CRF & LUKE \\\midrule
 		Model type & \multicolumn{3}{c}{BertForTokenClassification} &
 		with CRF layer & LukeForEntitySpanClassification \\
 		input\_size & 128 & 128 & 128 & 128 & 128\\
        classifier\_dropout & 0.2 & 0.2 & 0.2 & 0.2 & 0.0 \\
        embeddings &
        BERT\texorpdfstring{\textsubscript{base-cased}}{} & BERT\texorpdfstring{\textsubscript{large-cased}}{} & BERT\texorpdfstring{\textsubscript{base-cased}}{} & BERT\texorpdfstring{\textsubscript{large-cased}}{} & LUKE\texorpdfstring{\textsubscript{base}}{} \\
 		\midrule
 		Training parameters & & & & & \\
 		\midrule
 		max\_epochs & 20 & 30 & 25 & 30 & 30 \\
        learning\_rate & $2\cdot10^{-5}$ & $2\cdot10^{-5}$ & $2\cdot10^{-5}$ & $2\cdot10^{-5}$ & $2\cdot10^{-5}$ \\
        mini\_batch\_size & 16 & 16 & 16 & 16 & 16 \\
        weight\_decay & $0.01$ & $0.01$ & $0.01$ & $0.01$ & $0.0$\\
        seed & 42 & 42 & 42 & 42 & 42\\
        max\_entity\_length &-&-&-&-& 512\\
        max\_mention\_length & - & - & - & - & 5\\
 	\bottomrule
 	\end{tabular}
    \caption{Our baselines' information and customized parameters.}
    \label{tab:baseline-hyperparameters}
\end{table}

For the BERT models, we tested the impact of two hyperparameters: the learning rate and the number of training epochs. Regarding the first one, we found $2\cdot10^{-5}$ to be the optimal value. The optimal value for the second hyperparameter differed per model. Interestingly, the number of training epochs was a crucial hyperparameter for less frequent entities -- COLOR, TASTE, PURPOSE and PART.

\begin{table}[!htbp]
    \footnotesize
	\centering
	\begin{tabular}{cccccc}
		\toprule
		& \multicolumn{5}{c}{Number of training epochs} \\
		\cmidrule(r){2-6}
		entity & 10 & 15 & 20 & 25 & 30\\
		\midrule
		FOOD & $0.891 \pm 0.017$ & $0.893 \pm 0.016 $ & $0.896 \pm 0.014$ & $0.893 \pm 0.008$ & $\mathbf{0.898 \pm 0.012}$\\
		QUANTITY & $0.980 \pm 0.007$ & $0.982 \pm 0.007$ & $0.981 \pm 0.007$ & $\mathbf{0.983 \pm 0.005}$ & $0.982 \pm 0.007$\\ 
		UNIT & $0.978 \pm 0.008$ & $0.976 \pm 0.008$ & $0.977 \pm 0.007$ & $\mathbf{0.978 \pm 0.005}$ & $0.976 \pm 0.009$ \\ 
		PROCESS & $0.918 \pm 0.017$ & $0.913 \pm 0.019$ & $0.919 \pm 0.016$ & $0.921 \pm 0.015$ & $\mathbf{0.927 \pm 0.015}$\\
		PHYSICAL QUALITY & $0.787 \pm 0.025$ & $0.790 \pm 0.016$ & $0.783 \pm 0.019$ & $\mathbf{0.803 \pm 0.015}$ & $0.795 \pm 0.025$ \\
		COLOR & $0.879 \pm 0.032$ & $0.902 \pm 0.039$ & $0.905 \pm 0.032$ & $\mathbf{0.911 \pm 0.030}$ & $0.910 \pm 0.039$ \\
		TASTE & $0.752 \pm 0.139$ & $0.775 \pm 0.091$ & $\mathbf{0.798 \pm 0.088}$ & $0.790 \pm 0.103$ & $0.781 \pm 0.089$\\
		PURPOSE & $0.835 \pm 0.138$ & $0.823 \pm 0.114$ & $0.806 \pm 0.122$ & $0.837 \pm 0.145$ & $\mathbf{0.853 \pm 0.141}$\\
		PART & $0.666 \pm 0.100$ & $0.763 \pm 0.055$ & $0.739 \pm 0.084$ & $0.736 \pm 0.039$ & $\mathbf{0.801 \pm 0.034}$\\
		all & $0.930 \pm 0.012 $ & $0.931 \pm 0.013$ & $0.932 \pm 0.010$ & $0.934 \pm 0.008$ & $\mathbf{0.935 \pm 0.011}$\\
		\bottomrule
	\end{tabular}
	\caption{The impact on the training time on the BERT\texorpdfstring{\textsubscript{large-cased}}{} results (average $F_1$-scores over 5 runs in 5-fold cross-validation)
	\label{tab:bert-large-cased-num-epochs}}
\end{table}

Table~\ref{tab:bert-large-cased-num-epochs} illustrates the impact of the longer training, on BERT\texorpdfstring{\textsubscript{large-cased}}~~performance. While increasing the training time had subtle effect on the most popular entities, it substantially improved the performance on the rarest entities.

\section{Conclusions}\label{sec:conclusions}

We collected a novel dataset with a rich set of entities with crucial information about ingredients. We also tested several benchmarks based on Named Entity Recognition approaches to extract the critical information from recipes.

Our datasets and benchmarks have the potential to be helpful in, e.g.~determining similarity between recipes, understanding and recommending new recipes or ingredients based on deep knowledge of food products as ingredients and their dietary and functional effects, translating recipes between languages, generating new recipes and estimating nutrition values of ingredients in recipes.

Moreover, our dataset contains non-trivial challenges, such as rare cases of discontinuous annotations that may be challenging for NERs.


\section*{Acknowledgements}
The research leading to these results has received funding from the Norway Grants 2014-2021 via the National Centre for Research and Development. Project registration number: NOR/SGS/TAISTI/0323/2020.\\This research was also carried out with the support of the Laboratory of Bioinformatics and Computational Genomics and the High Performance Computing Center of the Faculty of Mathematics and Information Science Warsaw University of Technology under computational grant A-21-18.

\bibliographystyle{unsrtnat}
\bibliography{references}  

\begin{thebibliography}{25}
\providecommand{\natexlab}[1]{#1}
\providecommand{\url}[1]{\texttt{#1}}
\expandafter\ifx\csname urlstyle\endcsname\relax
  \providecommand{\doi}[1]{doi: #1}\else
  \providecommand{\doi}{doi: \begingroup \urlstyle{rm}\Url}\fi

\bibitem[Min et~al.(2019)Min, Jiang, Liu, Rui, and Jain]{food-computing-survey}
Weiqing Min, Shuqiang Jiang, Linhu Liu, Yong Rui, and Ramesh Jain.
\newblock A survey on food computing.
\newblock \emph{ACM Comput. Surv.}, 52\penalty0 (5), sep 2019.
\newblock ISSN 0360-0300.
\newblock \doi{10.1145/3329168}.
\newblock URL \url{https://doi.org/10.1145/3329168}.

\bibitem[Menichetti et~al.(2021)Menichetti, Ravandi, Mozaffarian, and
  Barab{\'a}si]{Menichetti2021.05.22.21257615}
Giulia Menichetti, Babak Ravandi, Dariush Mozaffarian, and
  Albert-L{\'a}szl{\'o} Barab{\'a}si.
\newblock Machine learning prediction of food processing.
\newblock \emph{medRxiv}, 2021.
\newblock \doi{10.1101/2021.05.22.21257615}.
\newblock URL
  \url{https://www.medrxiv.org/content/early/2021/05/25/2021.05.22.21257615}.

\bibitem[Marín et~al.(2021)Marín, Biswas, Ofli, Hynes, Salvador, Aytar,
  Weber, and Torralba]{recipe1m}
Javier Marín, Aritro Biswas, Ferda Ofli, Nicholas Hynes, Amaia Salvador, Yusuf
  Aytar, Ingmar Weber, and Antonio Torralba.
\newblock Recipe1m+: A dataset for learning cross-modal embeddings for cooking
  recipes and food images.
\newblock \emph{IEEE Transactions on Pattern Analysis and Machine
  Intelligence}, 43\penalty0 (1):\penalty0 187--203, 2021.
\newblock \doi{10.1109/TPAMI.2019.2927476}.

\bibitem[Bien et~al.(2020)Bien, Gilski, Maciejewska, Taisner, Wisniewski, and
  Lawrynowicz]{DBLP:conf/inlg/BienGMTWL20}
Michal Bien, Michal Gilski, Martyna Maciejewska, Wojciech Taisner, Dawid
  Wisniewski, and Agnieszka Lawrynowicz.
\newblock Recipe{NLG}: {A} cooking recipes dataset for semi-structured text
  generation.
\newblock In Brian Davis, Yvette Graham, John~D. Kelleher, and Yaji Sripada,
  editors, \emph{Proceedings of the 13th International Conference on Natural
  Language Generation, {INLG} 2020, Dublin, Ireland, December 15-18, 2020},
  pages 22--28. Association for Computational Linguistics, 2020.
\newblock URL \url{https://aclanthology.org/2020.inlg-1.4/}.

\bibitem[Mori et~al.(2014)Mori, Maeta, Yamakata, and Sasada]{mori2014flow}
Shinsuke Mori, Hirokuni Maeta, Yoko Yamakata, and Tetsuro Sasada.
\newblock Flow graph corpus from recipe texts.
\newblock In \emph{Proceedings of the Ninth International Conference on
  Language Resources and Evaluation (LREC'14)}, pages 2370--2377, 2014.

\bibitem[De~la Torre et~al.(2008)De~la Torre, Hodgins, Bargteil, Martin, Macey,
  Collado, and Beltran]{de2008guide}
Fernando De~la Torre, Jessica Hodgins, Adam Bargteil, Xavier Martin, Justin
  Macey, Alex Collado, and Pep Beltran.
\newblock Guide to the carnegie mellon university multimodal activity
  (cmu-mmac) database. robotics institute (2008), 2008.

\bibitem[Popovski et~al.(2019{\natexlab{a}})Popovski, Seljak, and
  Eftimov]{popovski_foodbase_2019}
Gorjan Popovski, Barbara~Koroušić Seljak, and Tome Eftimov.
\newblock {FoodBase} corpus: a new resource of annotated food entities.
\newblock \emph{Database}, 2019\penalty0 (baz121), January 2019{\natexlab{a}}.
\newblock ISSN 1758-0463.
\newblock \doi{10.1093/database/baz121}.
\newblock URL \url{https://doi.org/10.1093/database/baz121}.

\bibitem[Nasar et~al.(2021)Nasar, Jaffry, and
  Malik]{DBLP:journals/csur/NasarJM21}
Zara Nasar, Syed~Waqar Jaffry, and Muhammad~Kamran Malik.
\newblock Named entity recognition and relation extraction: State-of-the-art.
\newblock \emph{{ACM} Comput. Surv.}, 54\penalty0 (1):\penalty0 20:1--20:39,
  2021.
\newblock \doi{10.1145/3445965}.
\newblock URL \url{https://doi.org/10.1145/3445965}.

\bibitem[Lafferty et~al.(2001)Lafferty, McCallum, and
  Pereira]{DBLP:conf/icml/LaffertyMP01}
John~D. Lafferty, Andrew McCallum, and Fernando C.~N. Pereira.
\newblock Conditional random fields: Probabilistic models for segmenting and
  labeling sequence data.
\newblock In Carla~E. Brodley and Andrea~Pohoreckyj Danyluk, editors,
  \emph{Proceedings of the Eighteenth International Conference on Machine
  Learning {(ICML} 2001), Williams College, Williamstown, MA, USA, June 28 -
  July 1, 2001}, pages 282--289. Morgan Kaufmann, 2001.

\bibitem[Huang et~al.(2015)Huang, Xu, and Yu]{DBLP:journals/corr/HuangXY15}
Zhiheng Huang, Wei Xu, and Kai Yu.
\newblock Bidirectional {LSTM-CRF} models for sequence tagging.
\newblock \emph{CoRR}, abs/1508.01991, 2015.
\newblock URL \url{http://arxiv.org/abs/1508.01991}.

\bibitem[Hochreiter and Schmidhuber(1997)]{hochreiter1997long}
Sepp Hochreiter and J{\"u}rgen Schmidhuber.
\newblock Long short-term memory.
\newblock \emph{Neural computation}, 9\penalty0 (8):\penalty0 1735--1780, 1997.

\bibitem[Devlin et~al.(2019{\natexlab{a}})Devlin, Chang, Lee, and
  Toutanova]{devlin-etal-2019-bert}
Jacob Devlin, Ming-Wei Chang, Kenton Lee, and Kristina Toutanova.
\newblock {BERT}: Pre-training of deep bidirectional transformers for language
  understanding.
\newblock In \emph{Proceedings of the 2019 Conference of the North {A}merican
  Chapter of the Association for Computational Linguistics: Human Language
  Technologies, Volume 1 (Long and Short Papers)}, pages 4171--4186,
  Minneapolis, Minnesota, June 2019{\natexlab{a}}. Association for
  Computational Linguistics.
\newblock \doi{10.18653/v1/N19-1423}.
\newblock URL \url{https://aclanthology.org/N19-1423}.

\bibitem[Vaswani et~al.(2017)Vaswani, Shazeer, Parmar, Uszkoreit, Jones, Gomez,
  Kaiser, and Polosukhin]{DBLP:conf/nips/VaswaniSPUJGKP17}
Ashish Vaswani, Noam Shazeer, Niki Parmar, Jakob Uszkoreit, Llion Jones,
  Aidan~N. Gomez, Lukasz Kaiser, and Illia Polosukhin.
\newblock Attention is all you need.
\newblock In Isabelle Guyon, Ulrike von Luxburg, Samy Bengio, Hanna~M. Wallach,
  Rob Fergus, S.~V.~N. Vishwanathan, and Roman Garnett, editors, \emph{Advances
  in Neural Information Processing Systems 30: Annual Conference on Neural
  Information Processing Systems 2017, December 4-9, 2017, Long Beach, CA,
  {USA}}, pages 5998--6008, 2017.
\newblock URL
  \url{https://proceedings.neurips.cc/paper/2017/hash/3f5ee243547dee91fbd053c1c4a845aa-Abstract.html}.

\bibitem[Yamada et~al.(2020)Yamada, Asai, Shindo, Takeda, and
  Matsumoto]{yamada-etal-2020-luke}
Ikuya Yamada, Akari Asai, Hiroyuki Shindo, Hideaki Takeda, and Yuji Matsumoto.
\newblock {LUKE}: Deep contextualized entity representations with entity-aware
  self-attention.
\newblock In \emph{Proceedings of the 2020 Conference on Empirical Methods in
  Natural Language Processing (EMNLP)}, pages 6442--6454, Online, November
  2020. Association for Computational Linguistics.
\newblock \doi{10.18653/v1/2020.emnlp-main.523}.
\newblock URL \url{https://aclanthology.org/2020.emnlp-main.523}.

\bibitem[Akbik et~al.(2018)Akbik, Blythe, and
  Vollgraf]{DBLP:conf/coling/AkbikBV18}
Alan Akbik, Duncan Blythe, and Roland Vollgraf.
\newblock Contextual string embeddings for sequence labeling.
\newblock In Emily~M. Bender, Leon Derczynski, and Pierre Isabelle, editors,
  \emph{Proceedings of the 27th International Conference on Computational
  Linguistics, {COLING} 2018, Santa Fe, New Mexico, USA, August 20-26, 2018},
  pages 1638--1649. Association for Computational Linguistics, 2018.
\newblock URL \url{https://aclanthology.org/C18-1139/}.

\bibitem[Perera et~al.(2022)Perera, Nguyen, Dehmer, and
  Emmert-Streib]{make4010012}
Nadeesha Perera, Thi Thuy~Linh Nguyen, Matthias Dehmer, and Frank
  Emmert-Streib.
\newblock Comparison of text mining models for food and dietary constituent
  named-entity recognition.
\newblock \emph{Machine Learning and Knowledge Extraction}, 4\penalty0
  (1):\penalty0 254--275, 2022.
\newblock ISSN 2504-4990.
\newblock \doi{10.3390/make4010012}.
\newblock URL \url{https://www.mdpi.com/2504-4990/4/1/12}.

\bibitem[Cenikj et~al.(2020)Cenikj, Popovski, Stojanov, Seljak, and
  Eftimov]{cenikj_butter_2020}
Gjorgjina Cenikj, Gorjan Popovski, Riste Stojanov, Barbara~Koroušić Seljak,
  and Tome Eftimov.
\newblock {BuTTER}: {BidirecTional} {LSTM} for {Food} {Named}-{Entity}
  {Recognition}.
\newblock In \emph{2020 {IEEE} {International} {Conference} on {Big} {Data}
  ({Big} {Data})}, pages 3550--3556, December 2020.
\newblock \doi{10.1109/BigData50022.2020.9378151}.

\bibitem[Popovski et~al.(2019{\natexlab{b}})Popovski, Kochev, Seljak, and
  Eftimov]{popovski_foodie:_2019}
Gorjan Popovski, Stefan Kochev, Barbara Seljak, and Tome Eftimov.
\newblock {FoodIE}: {A} {Rule}-based {Named}-entity {Recognition} {Method} for
  {Food} {Information} {Extraction}:.
\newblock In \emph{Proceedings of the 8th {International} {Conference} on
  {Pattern} {Recognition} {Applications} and {Methods}}, pages 915--922,
  Prague, Czech Republic, 2019{\natexlab{b}}. SCITEPRESS - Science and
  Technology Publications.
\newblock ISBN 9789897583513.
\newblock \doi{10.5220/0007686309150922}.
\newblock URL
  \url{http://www.scitepress.org/DigitalLibrary/Link.aspx?doi=10.5220/0007686309150922}.

\bibitem[Jonquet et~al.(2009)Jonquet, Shah, Youn, Callendar, Storey, and
  Musen]{jonquet2009ncbo}
Clement Jonquet, Nigam Shah, Cherie Youn, Chris Callendar, Margaret-Anne
  Storey, and M~Musen.
\newblock Ncbo annotator: semantic annotation of biomedical data.
\newblock In \emph{International Semantic Web Conference, Poster and Demo
  session}, volume 110. Washington DC, USA, 2009.

\bibitem[Popovski et~al.(2020)Popovski, Seljak, and
  Eftimov]{popovski_survey_2020}
Gorjan Popovski, Barbara~Koroušić Seljak, and Tome Eftimov.
\newblock A {Survey} of {Named}-{Entity} {Recognition} {Methods} for {Food}
  {Information} {Extraction}.
\newblock \emph{IEEE Access}, 8:\penalty0 31586--31594, 2020.
\newblock ISSN 2169-3536.
\newblock \doi{10.1109/ACCESS.2020.2973502}.

\bibitem[Stenetorp et~al.(2012)Stenetorp, Pyysalo, Topi{\'c}, Ohta, Ananiadou,
  and Tsujii]{stenetorp-etal-2012-brat}
Pontus Stenetorp, Sampo Pyysalo, Goran Topi{\'c}, Tomoko Ohta, Sophia
  Ananiadou, and Jun{'}ichi Tsujii.
\newblock brat: a web-based tool for {NLP}-assisted text annotation.
\newblock In \emph{Proceedings of the Demonstrations at the 13th Conference of
  the {E}uropean Chapter of the Association for Computational Linguistics},
  pages 102--107, Avignon, France, April 2012. Association for Computational
  Linguistics.
\newblock URL \url{https://aclanthology.org/E12-2021}.

\bibitem[Devlin et~al.(2019{\natexlab{b}})Devlin, Chang, Lee, and
  Toutanova]{devlin_bert:_2019}
Jacob Devlin, Ming-Wei Chang, Kenton Lee, and Kristina Toutanova.
\newblock {BERT}: {Pre}-training of {Deep} {Bidirectional} {Transformers} for
  {Language} {Understanding}.
\newblock \emph{arXiv:1810.04805 [cs]}, May 2019{\natexlab{b}}.
\newblock URL \url{http://arxiv.org/abs/1810.04805}.
\newblock arXiv: 1810.04805.

\bibitem[Stojanov et~al.(2021)Stojanov, Popovski, Cenikj, Seljak, and
  Eftimov]{stojanov_fine-tuned_2021}
Riste Stojanov, Gorjan Popovski, Gjorgjina Cenikj, Barbara~Koroušić Seljak,
  and Tome Eftimov.
\newblock A {Fine}-{Tuned} {Bidirectional} {Encoder} {Representations} {From}
  {Transformers} {Model} for {Food} {Named}-{Entity} {Recognition}: {Algorithm}
  {Development} and {Validation}.
\newblock \emph{Journal of Medical Internet Research}, 23\penalty0
  (8):\penalty0 e28229, August 2021.
\newblock \doi{10.2196/28229}.
\newblock URL \url{https://www.jmir.org/2021/8/e28229}.

\bibitem[Souza et~al.(2020)Souza, Nogueira, and Lotufo]{souza2020bertimbau}
F{\'a}bio Souza, Rodrigo Nogueira, and Roberto Lotufo.
\newblock {BERT}imbau: pretrained {BERT} models for {B}razilian {P}ortuguese.
\newblock In \emph{9th Brazilian Conference on Intelligent Systems, {BRACIS},
  Rio Grande do Sul, Brazil, October 20-23 (to appear)}, 2020.

\bibitem[Segura-Bedmar et~al.(2013)Segura-Bedmar, Mart{\'\i}nez, and
  Herrero-Zazo]{segura-bedmar-etal-2013-semeval}
Isabel Segura-Bedmar, Paloma Mart{\'\i}nez, and Mar{\'\i}a Herrero-Zazo.
\newblock {S}em{E}val-2013 task 9 : Extraction of drug-drug interactions from
  biomedical texts ({DDIE}xtraction 2013).
\newblock In \emph{Second Joint Conference on Lexical and Computational
  Semantics (*{SEM}), Volume 2: Proceedings of the Seventh International
  Workshop on Semantic Evaluation ({S}em{E}val 2013)}, pages 341--350, Atlanta,
  Georgia, USA, June 2013. Association for Computational Linguistics.
\newblock URL \url{https://aclanthology.org/S13-2056}.

\end{thebibliography}

\appendix
\section{Detailed Dataset Analysis}\label{sec:eda2}

\begin{figure*}[h!]
\centering
\includegraphics[width=.7\textwidth]{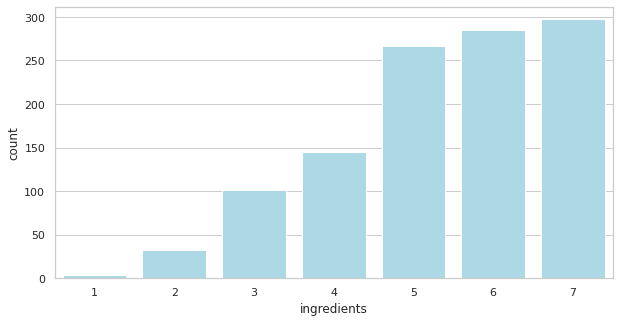}\hfill
\caption{The distribution of the number of ingredients}
\label{fig:ingredient_number_dist}
\end{figure*}

Analysis of the distribution of the number of ingredients in the dataset reveal the recipes' diversity -- see Figure~\ref{fig:ingredient_number_dist}.

\begin{figure*}[h!]
\centering
\includegraphics[width=.7\textwidth]{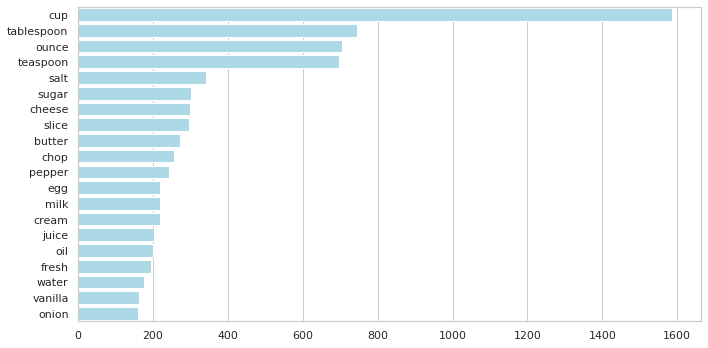}\hfill
\caption{Most frequent non stop-words}
\label{fig:most_frequent_non_stopwords}
\end{figure*}

The most common non-stop words are shown in Figure~\ref{fig:most_frequent_non_stopwords}

\begin{figure*}[h!]
\centering
\includegraphics[width=.7\textwidth]{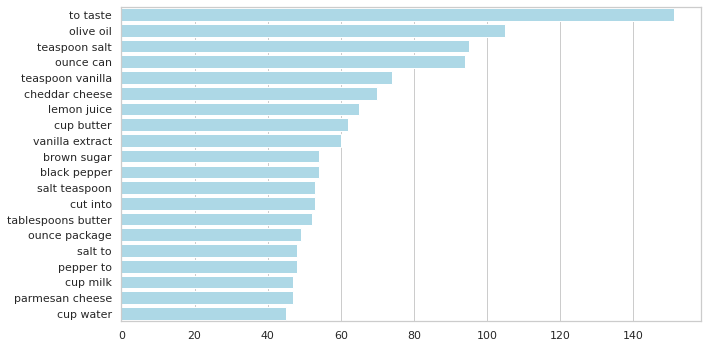}\hfill
\caption{Most frequent bigrams}
\label{fig:most_frequent_bigrams}
\end{figure*}

The most frequent bigrams are shown in Figure~\ref{fig:most_frequent_bigrams}.

\begin{figure*}[h!]
\centering
\includegraphics[width=.7\textwidth]{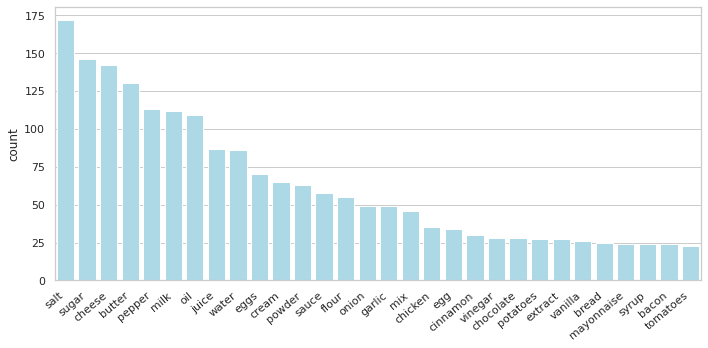}\hfill
\caption{Most frequent words in the food entity}
\label{fig:food_word_distribution}
\end{figure*}

The most frequent words from the entity food are shown in Figure~\ref{fig:food_word_distribution}.

\begin{figure*}[h!]
\centering
\includegraphics[width=.7\textwidth]{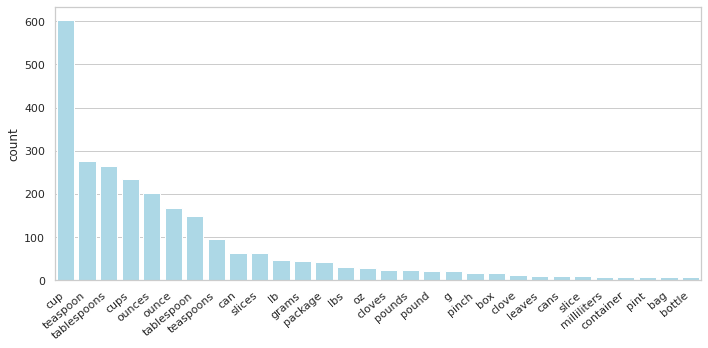}\hfill
\caption{Most frequent words in the unit entity}
\label{fig:unit_word_distribution}
\end{figure*}

The most frequent words from the entity unit are shown in Figure~\ref{fig:unit_word_distribution}.

\begin{figure*}[h!]
\centering
\includegraphics[width=.7\textwidth]{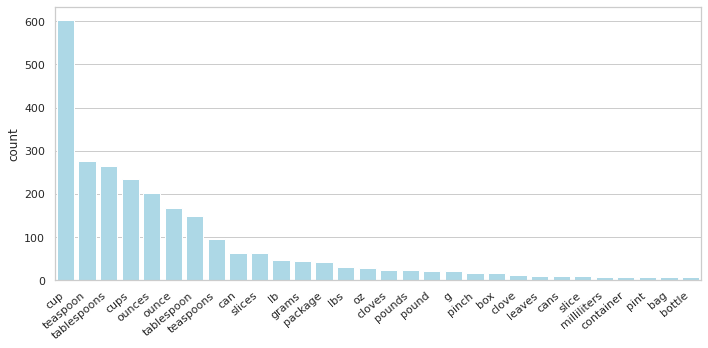}\hfill
\caption{Most frequent words in the quality entity}
\label{fig:quality_word_distribution}
\end{figure*}

The most frequent words from the entity quality are shown in Figure~\ref{fig:quality_word_distribution}.

\begin{figure*}[h!]
\centering
\includegraphics[width=.7\textwidth]{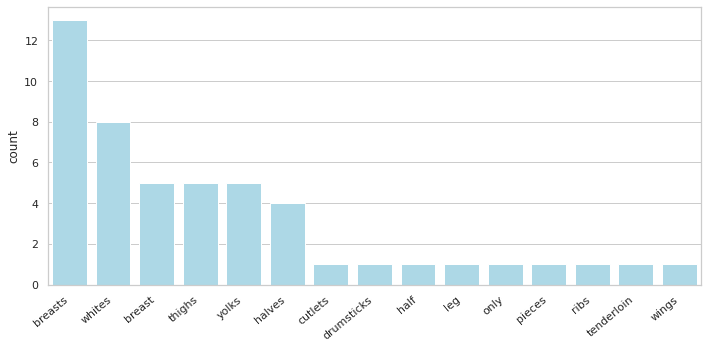}\hfill
\caption{Most frequent words in the part entity}
\label{fig:part_word_distribution}
\end{figure*}

The most frequent words from the entity part are shown in Figure~\ref{fig:part_word_distribution}.

\begin{figure*}[h!]
\centering
\includegraphics[width=.7\textwidth]{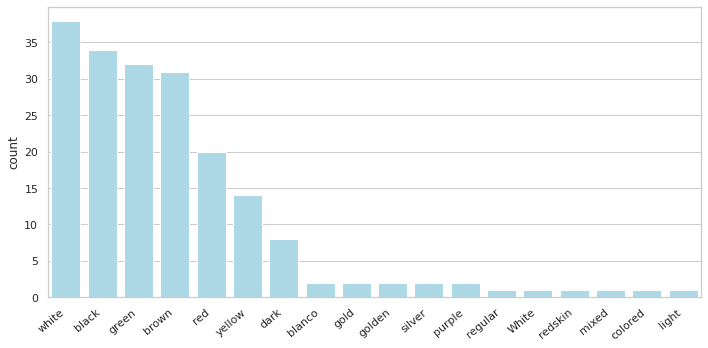}\hfill
\caption{Most frequent words in the color entity}
\label{fig:color_word_distribution}
\end{figure*}

The most frequent words from the entity color are shown in Figure~\ref{fig:color_word_distribution}.

\begin{figure*}[h!]
\centering
\includegraphics[width=.7\textwidth]{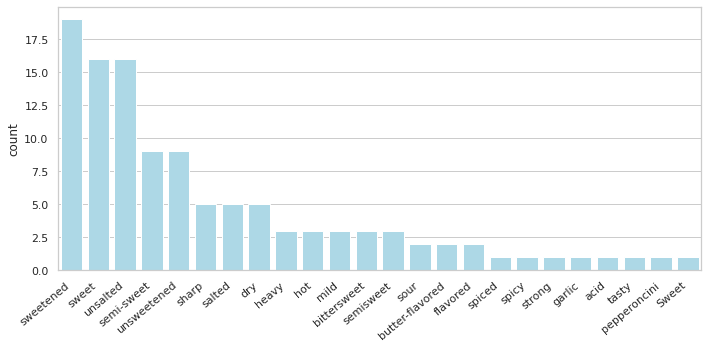}\hfill
\caption{Most frequent words in the taste entity}
\label{fig:taste_word_distribution}
\end{figure*}

The most frequent words from the entity taste are shown in Figure~\ref{fig:taste_word_distribution}.

\begin{figure*}[h!]
\centering
\includegraphics[width=.7\textwidth]{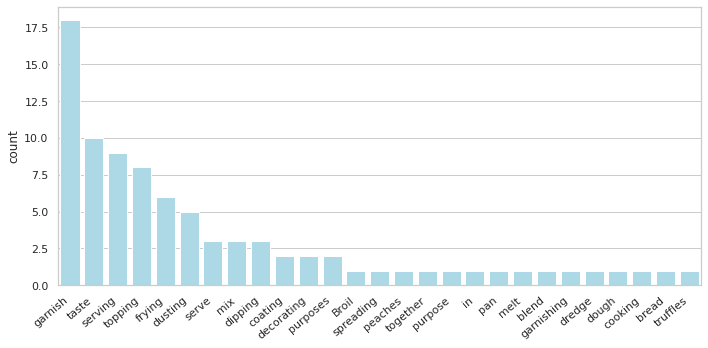}\hfill
\caption{Most frequent words in the purpose entity}
\label{fig:purpose_word_distribution}
\end{figure*}

The most frequent words from the entity purpose are shown in Figure~\ref{fig:purpose_word_distribution}.

\end{document}